\documentclass[10pt]{article}

\usepackage{amsmath}
\usepackage{amssymb}

\usepackage{graphicx}

\usepackage{cite}
\usepackage{bm,bbm}
\usepackage{booktabs}
\usepackage{color} 

\usepackage[labelfont=bf,labelsep=period,justification=raggedright]{caption}

\makeatletter
\renewcommand{\@biblabel}[1]{\quad#1.}
\makeatother

\date{today}

\begin{document}

% Title must be 150 characters or less
\begin{flushleft}
{\Large
\textbf{Distinguishing noise from chaos:  objective versus subjective criteria using Horizontal Visibility Graph.}
}
% Insert Author names, affiliations and corresponding author email.

\vspace{0.3cm}
Mart\'in G\'omez Ravetti  $^{1}$, 
Laura C. Carpi $^{2}$,
Bruna Amin Gon\c calves$^{1}$,
Alejandro C. Frery $^{2}$,
Osvaldo A. Rosso $^{3}$ 

\vspace{0.3cm}
\bf{1} Departamento de Engenharia de Produ\c{c}\~ao, Universidade Federal de Minas Gerais. \\
		Av. Ant\^onio Carlos, 6627, Belo Horizonte, 31270-901 Belo Horizonte - MG, Brazil.
\\
\bf{2} LaCCAN/CPMAT - Instituto de Computa\c{c}\~ao, Universidade Federal de Alagoas \\
		BR 104 Norte km 97, 57072-970 Macei\'o, Alagoas - Brazil.
\\
\bf{3} Instituto de F\'{\i}sica, Universidade Federal de Alagoas. \\
		BR 104 Norte km 97, 57072-970 Macei\'o, Alagoas - Brazil.
\\
		Laboratorio de Sistemas Complejos, Facultad de Ingenier\'{\i}a,\\
		Universidad de Buenos Aires (UBA).\\
		(1063) Av. Paseo Col\'on 840, Ciudad Aut\'onoma de Buenos Aires, Argentina.
\\
$\ast$ E-mail: martin.ravetti@dep.ufmg.br

\end{flushleft}

% Please keep the abstract between 250 and 300 words
\section*{Abstract}
A recently proposed methodology called the Horizontal Visibility Graph (HVG) 
[Luque {\it et al.}, Phys. Rev. E., 80, 046103 (2009)] that constitutes a geometrical simplification of  
the well known Visibility Graph algorithm [Lacasa {\it et al.\/}, Proc. Natl. Sci. U.S.A. 105, 4972 (2008)], 
has been used to study the distinction between deterministic and stochastic components in time series 
[L. Lacasa and R. Toral, Phys. Rev. E., 82, 036120 (2010)]. 
Specifically, the authors propose that the node degree distribution of these processes follows an 
exponential functional of the form $P(\kappa)\sim \exp(-\lambda~\kappa)$, in which $\kappa$ is the node 
degree and $\lambda$ is a positive parameter able to distinguish between deterministic (chaotic) and 
stochastic (uncorrelated and correlated) dynamics. 
In this work, we investigate the characteristics of the node degree distributions constructed by using HVG, 
for time series corresponding to $28$ chaotic maps and $3$ different stochastic processes.
We thoroughly study the methodology proposed by Lacasa and Toral finding several cases for which their 
hypothesis is not valid.  
We propose a methodology that uses the HVG together with Information Theory quantifiers.  
An extensive and careful analysis of the node degree distributions obtained by applying HVG 
allow us to conclude that the Fisher-Shannon information plane is  a remarkable tool able to graphically represent 
the different nature, deterministic or stochastic, of the systems under study. 

%\section*{Author Summary}

\section{Introduction}
\label{sec:Intro}     

The distinction between stochastic and chaotic processes has received much attention, becoming one 
of the most appealing problems in time series analysis. 
Since stochastic and chaotic (low dimensional deterministic) processes share several characteristics, 
the discrimination between them is a challenging task. 
Time series with complex structures are very frequent in both natural and artificial systems. 
The interest behind this distinction relies in uncovering the cause of unpredictability governing these systems. 

Much effort has being dedicated in the understanding of this topic. It was thought, in the origins of 
chaotic dynamics, 
that obtaining finite, non-integer values for fractal dimension was a strong evidence of the presence of 
deterministic chaos, as stochastic processes were thought to have an infinite value.  
Osborne and Provenzale \cite{Osborne1989} observed for a stochastic process a non-convergence 
in the correlation dimension (as a estimation of fractal dimension). They showed that time series 
generated by inverting power law spectra and random phases 
are random fractal paths with finite Hausdorff dimension and, consequently, with finite correlation dimension 
\cite{Osborne1989}.

Among other methodologies to distinguish chaotic from stochastic time series we can mention the work of 
Sugihara and May \cite{Sugihara1990} based on nonlinear forecasting in which they compare predicted and actual 
trajectories and make tentative distinctions between dynamical chaos and measurement errors. 
The accuracy of nonlinear forecast diminishes for increasing prediction time-intervals for a chaotic time series. 
This dependency is not found for uncorrelated noises \cite{Sugihara1990}. 
Kaplan and Glass \cite{Kaplan1992,Kaplan1993} observed that the tangent to the trajectory generated by a 
deterministic system is a function of the position in phase space, consequently, all the tangents to a 
trajectory in a given phase space region will display similar orientation. 
As stochastic dynamics do not exhibit this behavior, Kaplan and Glass proposed a test based on these observations.
Kantz and co-workers \cite{Kantz2000,Cencini2000} recently analyzed the behavior of entropy quantifiers as a 
function of the coarse-graining resolution, and applied their ideas to distinguish between chaos and noise. 
Their methodology can be considered a generalization of the Grassberger and Procaccia method
\cite{Grassberger1983} regarding the estimation of the correlation dimension and the consideration of finite values as signatures of deterministic behavior. 

Chaotic systems display ``sensitivity to initial conditions" which manifests instability everywhere 
in the phase space and leads to non-periodic motion (chaotic time series). 
They display long-term unpredictability despite the deterministic character of the temporal trajectory. 
In a system undergoing chaotic motion, two neighboring points in the phase space move away exponentially rapidly. 
Let ${\mathbf x}_1(t)$ and ${\mathbf x}_2(t)$ be two such points, located within a ball of radius $R$ at time $t$. 
Further, assume that these two points cannot be resolved within the ball due to poor instrumental resolution. 
At some later time $t'$ the distance between the points will typically grow to 
$|{\mathbf x}_1(t')-{\mathbf x}_2(t')| \approx |{\mathbf x}_1(t)-{\mathbf x}_2(t)|~\exp( \Lambda~|t'-t|)$, with 
$\Lambda > 0$ for a chaotic dynamics, being $\Lambda$ the biggest Lyapunov exponent. 
When this distance at time $t'$ exceeds $R$, the points become experimentally distinguishable. 
This  implies that instability reveals some information about the phase space population that was not available 
at earlier times \cite{Abarbanel1996}. 
The above considerations allow to think chaos as an {\it information source}.
Moreover, the associated rate of generated information can be formulated in a precise way in terms of 
Kolmogorov-Sinai's entropy \cite{Kolmogorov1958,Sinai1959}.

In more recent works, the use of quantifiers based on Information Theory has led to interesting results 
regarding the characteristics of nonlinear chaotic dynamics, improving the understanding of their associated 
time series. Specifically, Rosso and collaborators \cite{Rosso2007,Rosso2011,Rosso2012,Rosso2013} found that
the use of the normalized Shannon entropy and statistical complexity allow for a better distinction between 
stochastic and chaotic dynamics, when incorporating causal information via the Bandt and Pompe  methodology 
\cite{Bandt2002,Zanin2012}, generating in this way the graphic tool called {\it the causality entropy-complexity plane}. 
In the same fashion, Olivares {\it et al.\/} \cite{Olivares2012A,Olivares2012B} propose the use of two information 
quantifiers like measures, namely, the normalized Shannon entropy and the Fisher information combined in the 
so-called {\it the causality Shannon-Fisher plane}, finding that the two dynamics exhibit different planar locations. 

The use of Visibility Graphs (VG) introduced by Lacasa and co-workers \cite{Lacasa2008}, a method that transform 
a time series into a graph, has also been used with this purpose.
Specifically, HVGs, a geometrical simplification of VGs, which is also 
computationally faster, was applied in the classification and characterization of periodic, chaotic, and onset of 
chaos dynamics \cite{Luque2011,Luque2012}. 
%Note that this methodology incorporates in a natural way the time causality present in the time series.

Lacasa and Toral \cite{Lacasa2010} studied the discrimination between chaotic, uncorrelated and correlated stochastic 
time series by using HVG.  They conjecture that the node degree distribution of these systems follows an exponential functional of the form $P(\kappa) \sim \exp(-\lambda~\kappa)$,  
in which $\lambda$ is a positive parameter and $\kappa$ the node degree. They computed analytically the HVG-PDF for the case of uncorrelated noise (white noise) \cite{Luque2009}, and found the
corresponding parameter value $\lambda_c = \ln( 3/2 )$.
Moreover, they hypothesized that this value corresponds to a central value that separates correlated stochastic ($\lambda > \lambda_c$) 
from chaotic dynamics ($\lambda < \lambda_c$). 

Even though the methodology works for several chaotic and stochastic systems, we have found several examples 
for which results diverge from the ones expected.

In this work, we present a methodology able to discriminate between chaotic and stochastic (uncorrelated and correlated) 
time series by using the HVG methodology together with Information Theory quantifiers. 
A total of $31$ systems are considered; the $27$ chaotic maps described by Sprott \cite{Sprott2003} and the Schuster map \cite{Schuster1988},  
noises with $f^{-k}$, $k \geq 0$ power spectrum (PS) and stochastic time series generated by fractional Brownian motion (fBm) and fractional Gaussian noise (fGn). 

Following Olivares {\it et al.\/} \cite{Olivares2012A,Olivares2012B} we based our analysis on the so-called 
Shannon-Fisher information plane (${\mathcal S} \times {\mathcal F}$) that captures both global and local features 
of the system's dynamics. 
Its horizontal and vertical axis are functionals of the pertinent probability distribution, namely, the normalized Shannon entropy (${\mathcal S}$) and the normalized Fisher Information measure (${\mathcal F}$). 
We evaluate  these quantifiers for the time series using as PDF the node degree distribution obtained via the horizontal visibility graph.
We show that the Shannon-Fisher information plane is able to efficiently represent the different nature of the systems in a planar representation, as well as to distinguish between the different degrees of correlation structures.

As for the organization of this work, the forthcoming Section \ref{sec:Chaos-Ruidos} enumerates and describes the chaotic maps 
and the stochastic processes considered. Section \ref{sec:HVG} describes the Horizontal Visibility Graph algorithm and discusses the characterization of the HVG-PDF from a statistical point of view, as well as the methodology implemented in \cite{Lacasa2010} based on the parameter $\lambda$. 
In Section \ref{sec:Shannon-Fisher} the basis on the Shannon-Fisher plane is detailed, and finally, Section \ref{sec:Results} present our results and discussions.

\section{Materials and Methods}
\subsection{Chaotic maps and stochastic processes}
\label{sec:Chaos-Ruidos}

\subsubsection{Chaotic maps}
\label{sec:Chaos}

 In the present work we consider $27$ chaotic maps described by Sprott in his book~\cite{Sprott2003} and the Schuster Maps~\cite{Schuster1988} (see supplementary material), grouped as follows:
 
\begin{description}
	\item [noninvertible maps:]
 
(1)~Logistic map~\cite{May1976};
(2)~Sine map~\cite{Strogatz1994};
(3)~Tent map~\cite{Devaney1989};
(4)~Linear congruential generator~\cite{Knuth1997};
(5)~Cubic map~\cite{Zeng1985};
(6)~Ricker's population model~\cite{Ricker1954};
(7)~Gauss map~\cite{vanWyk1997};
(8)~Cusp map~\cite{Beck1995};
(9)~Pinchers map~\cite{Potapov2000};
(10)~Spence map~\cite{Shaw1981};
(11)~Sine-circle map~\cite{Arnold1965}.

 	\item [dissipative maps:]
  
(12)~H\'enon map~\cite{Henon1976};
(13)~Lozi map~\cite{Lozi1978};
(14)~Delayed logistic map~\cite{Aronson1982};
(15)~Tinkerbell map~\cite{Nusse1994};
(16)~Burgers' map~\cite{Whitehead1984};
(17)~Holmes cubic map~\cite{Holmes1979};
(18)~Dissipative standard map~\cite{Schmidt1985};
(19)~Ikeda map~\cite{Ikeda1979};
(20)~Sinai map~\cite{Sinai1972};
(21)~Discrete predator-prey map~\cite{Beddington1975}. 

	 \item [conservative maps:]
  
(22)~Chirikov standard map~\cite{Chirikov1979};
(23)~H\'enon area-preserving quadratic map~\cite{Henon1969};
(24)~Arnold's cat map~\cite{Arnold1968};
(25)~Gingerbreadman map~\cite{Devaney1984};
(26)~Chaotic web map~\cite{Chernikov1988};
(27)~Lorenz three-dimensional chaotic map~\cite{Lorenz1993}.
\end{description}

We also analyze the Schuster map, a class introduced by Schuster and co-workers~\cite{Schuster1988} that exhibits intermittent signals with chaotic bursts and $f^{-z}$ power spectrum (PS).
It is defined as:

\begin{equation}
\label{eq:schuster} 
x_{n+1}~=~(x_n + x_n^z) \ \bmod{1} \ .
\end{equation}
 
As the parameter $z$ increases, the laminar zone increases in size and the chaotic bursts are less frequent. 
To generate these maps we use random initial conditions in the interval $(0,1)$ and we consider $z \in\{1.25, 1.50, 1.75, 2.00\}$.
 
For  noninvertible, dissipative and conservative maps  we use the initial conditions and parameter-values detailed by Sprott.
The corresponding initial values are given in the basin of attraction for noninvertible maps, near the attractor for dissipative 
maps, and in the chaotic sea for conservative maps \cite{Sprott2003}.
In the generation process $N$ iterations were considered after discarding the first $10^5$. 
In the case of multi-dimensional maps, we consider all map coordinates. 
A complete description of these maps can be found in the supplementary material.

\subsubsection{Stochastic processes}
\label{sec:Ruidos}
  
The following classical stochastic processes are considered in this work:

%\begin{itemize}
%	\item {\it Noises with $f^{-k}$ power spectrum generated as follows: \/}
\begin{description}
	\item [Noises with $f^{-k}$ power spectrum generated as follows:]
 
1) By using the Mersenne twister generator \cite{Mersenne1998}, the $\textsc{Matlab}^\copyright$ {\sl RAND\/}  function is used to produce pseudo random numbers in the interval $(-0.5,0.5)$ with an almost flat power spectrum (PS), uniform PDF, and zero mean value.

2) The Fast Fourier Transform (FFT) ${y^1_k}$ of the time series is obtained and multiplied by $f^{-k/2}$ 
($k>0$), yielding ${y^2_k}$; 

3) ${y^2_k}$ is symmetrized so as to obtain a real function. 
The pertinent inverse FFT is obtained after rounding off and truncation.
The ensuing time series ${x_i}$ has the desired power-spectrum properties and, by construction, is representative of non-Gaussian noises. In this work we consider noises in the range $0\leq k \leq 2.5$, with $\Delta k = 0.25$.  

%	\item {\it Fractional Brownian motion (fBm) and fractional Gaussian noise (fGn).\/} 
	\item [Fractional Brownian motion (fBm) and fractional Gaussian noise (fGn):] 
fBm is the only family of processes which is Gaussian, self-similar, and endowed with stationary increments (see Ref. \cite{Zunino2007} and references therein).
The normalized family of these Gaussian processes, $\{B^\mathcal{H}(t), t>0\}$, has the following properties:
{\it i)\/} $B^\mathcal{H}(0)=0$ almost surely,
{\it ii)\/} $\mathbb{E}[B^\mathcal{H}(t)]=0$ (zero mean), and
{\it iii)\/} covariance given by
\begin{equation}
\label{fbm-cov}
{\mathbb E} [{B^\mathcal{H}(t)
B^\mathcal{H}(s)}]~=~(t^{2\mathcal{H}}+s^{2\mathcal{H}}-|{t-s}|^{2\mathcal{H}})~/~2 \ ,
\end{equation}
for $s,t \in \mathbb{R}$.
Here ${\mathbb E}[\cdot ]$ refers to the mean.
The power exponent $0<\mathcal{H}<1$ is commonly known as the Hurst parameter or Hurst exponent.
These processes exhibit {\it memory} for any Hurst parameter except for $\mathcal{H}=1/2,$ as one realizes  
from Eq.~(\ref{fbm-cov}).
The $\mathcal{H}=1/2$ case corresponds to classical Brownian motion and  successive motion-increments are as likely 
to have the same sign as the opposite (there is no correlation among them).
Thus, Hurst's parameter defines two distinct regions in the interval $(0,1)$.
When $\mathcal{H}>1/2$, consecutive increments tend to have the same sign so that these processes are 
\textit{persistent}. On the contrary, for $\mathcal{H}<1/2$, consecutive increments are more likely to have opposite signs \textit{anti-persistent}.

Let us introduce the quantity $\{W^\mathcal{H}(t), t>0\}$ (fBm-``increments")
\begin{equation}
W^\mathcal{H}(t)~=~B^\mathcal{H}(t+1)-B^\mathcal{H}(t) \ ,
\label{eq:fgn}
\end{equation}
so as to express our Gaussian noise in the fashion

\begin{eqnarray}
 \rho(k) =& &{\mathbb E}[{W^\mathcal{H}(t)W^\mathcal{H}(t+k)}]\nonumber \\
            =& \frac{1}{2}&\left[(k+1)^{2\mathcal{H}} - 2k^{2\mathcal{H}}+|{k-1}|^{2\mathcal{H}}\right],~k>0 \ . 
\label{eq:fgn-cov}
\end{eqnarray}

Note that for $\mathcal{H}=1/2$ all correlations at nonzero lags vanish and $\{W^{1/2}(t), t>0\}$  represents \textit{white Gaussian noise}.
%\end{itemize}
\end{description}
 
The fBm and fGn are continuous but non-differentiable
processes (in the classical sense).
As non-stationary processes, they do  not possess a spectrum defined in the usual sense;
however, it is possible to define a \textit{generalized power spectrum} of
the form:
\begin{equation}
\Phi~\propto~ {\left|f\right|^{-\theta}} \ ,
\label{eq:fbn-fgn-PS}
\end{equation}
with
$\theta \equiv \alpha =2\mathcal{H}+1$, and $1<\alpha<3$ for fBm and;
$\theta \equiv \beta  =2\mathcal{H}-1$, and $-1<\beta<1,$ for fGn.

We use the Matlab function ``wfbm" that returns a fractional Brownian motion signal with a Hurst parameter $\mathcal{H}$ $(0 < \mathcal{H} < 1)$ and length $N$ for the generation the fBm and fGn time series. The algorithm was proposed by Abry and Sellan~\cite{Abry1996,Bardet2003}. 
In this work we consider noises in the range $0.1 \leq \mathcal{H} \leq 0.9$.

\subsection{Horizontal Visibility Graph}
\label{sec:HVG}
The horizontal visibility graph (HVG) \cite{Luque2009} is a geometrical simplification of the 
visibility graph (VG) \cite{Lacasa2008} that maintains the inherent characteristics of the transformed 
time series and incorporates in a natural way its time causality.

By construction, the HVG transforms a time series into a graph, 
in which each node corresponds to a point in the time series, will be connected considering the following criterion: 
\begin{quote}
	\item  
Let  $\{x_i, i=1, \dots, N\}$, be a time series of $N$ data. Two nodes $i$ and $j$ in the graph are connected if it is possible 
to trace a horizontal line in the time series linking $x_i$ and $x_j$ not intersecting intermediate data height, fulfilling: 
$x_i,~x_j~> x_n$ for all $i < n <  j$. 
\end{quote}

\begin{figure} [ht!]
\centering
\includegraphics[width=7in]{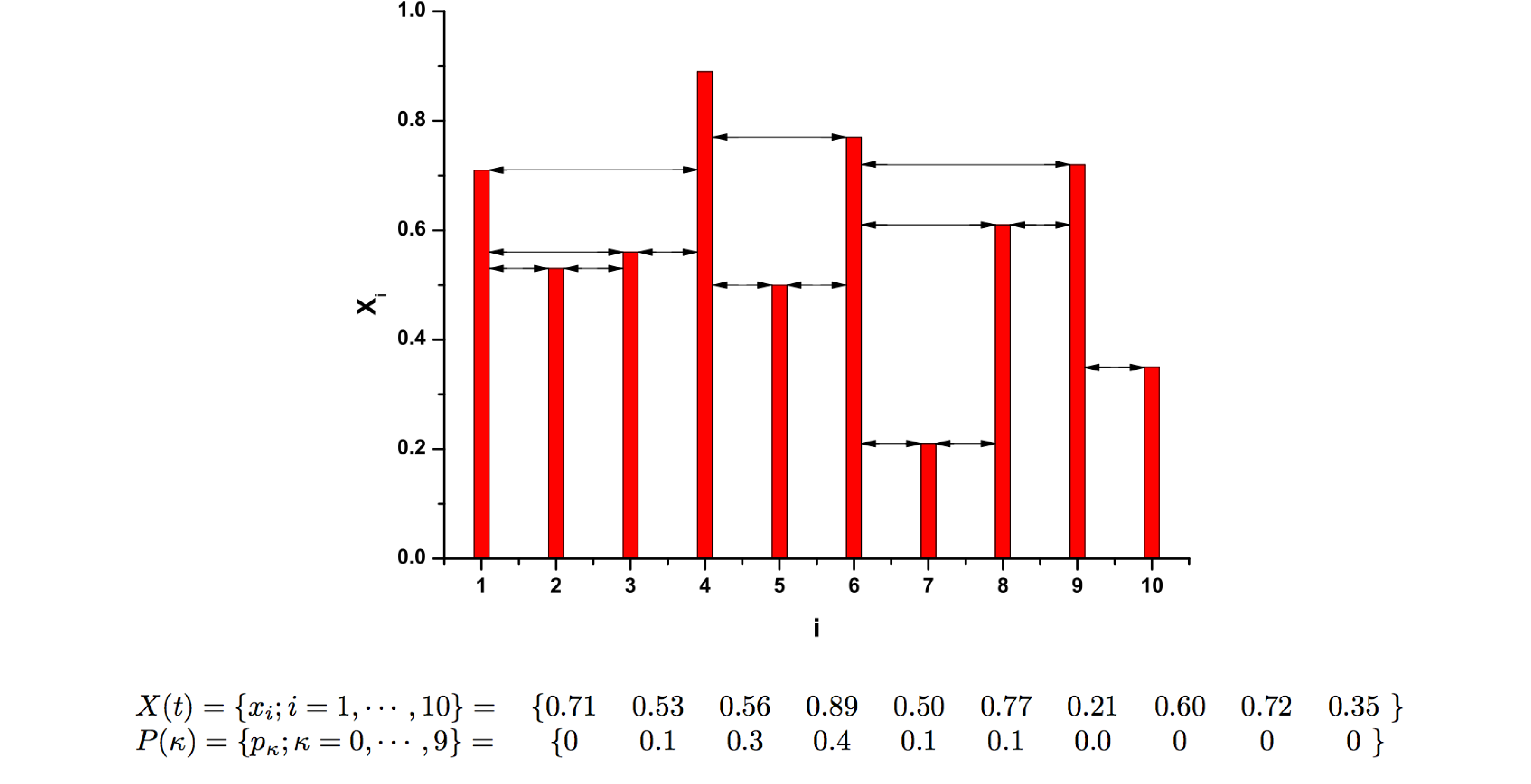}
\caption{ 
\label{fig:HVG-esquema} 
Horizontal Visibility Graph method applied to the time series $X(t)$.  
$P(\kappa)$ denotes the node degree distribution of the obtained graph (HVG-PDF).
}
\end{figure}

Note that the HVG preserves the time causality of the original series where each node sees at least its nearest neighbors.  
Another important feature of the HVG is the invariance under affine transformations, as its visibility is not modified under 
rescaling of horizontal and vertical axes, as well as under horizontal and vertical translations.  
Some other interesting properties are discussed in \cite{Luque2009}. An example of a time series 
and its associated node degree distribution based on HVG is given in Figure~\ref{fig:HVG-esquema}.

\subsubsection{The $\lambda$ rule}
\label{sec:LambdaRule}

Lacassa and Toral \cite{Lacasa2010} propose that chaotic and stochastic time series map into a graph with an exponential 
node degree distribution $P(\kappa)\sim \exp(-\lambda\kappa)$. 
The $\lambda$ parameter is computed by adjusting, using the least square method, a straight line being $\lambda$ its slope. 
The linear scaling region considered by Lacassa and Toral is  $3 \leq \kappa \leq 20$ or $3 \leq \kappa \leq \kappa_{\max}$ 
(if $\kappa_{\max} \leq 20$) for stochastic processes; and $3 \leq \kappa \leq 25$ or $3 \leq \kappa \leq \kappa_{\max}$ 
(if $\kappa_{\max} \leq 25$) for chaotic ones. 
The parameter $\lambda$ characterizes chaotic processes  when $\lambda < \ln(3/2)$, uncorrelated noises for $\lambda = \ln(3/2)$ 
and correlated noises when $\lambda > \ln(3/2)$. 

\begin{figure} [ht!]
\centering
\includegraphics[width=6.5in]{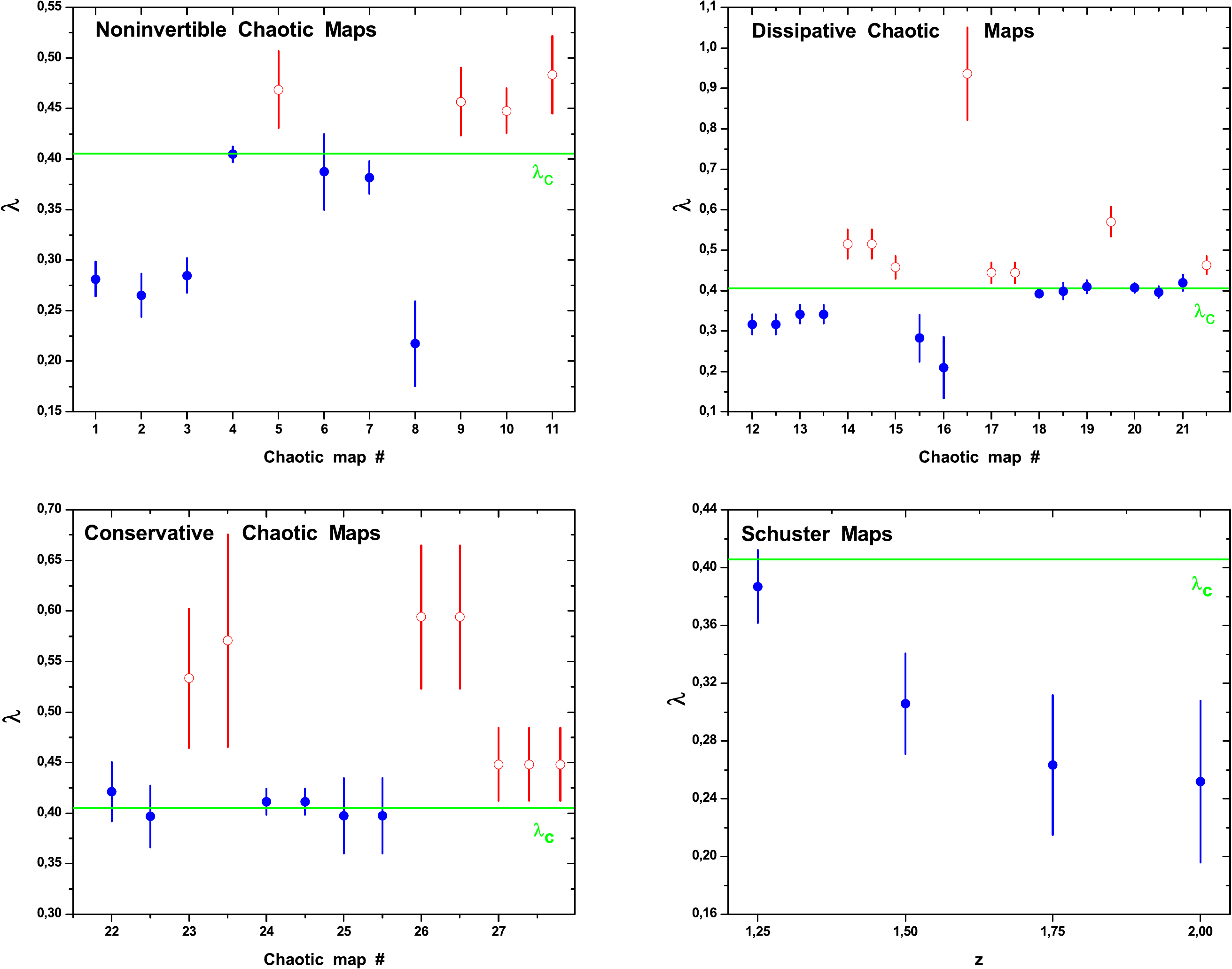}
\caption{
\label{fig:HVG-lambda-int-1} 
Confidence intervals for $\lambda$ values for Noninvertible, Dissipative,
Conservative and Schuster Chaotic Maps.  The values were obtained following the methodology proposed by Lacasa {\it et al.}. 
Symmetric confidence intervals at the $95\%$ confidence level were obtained for the $\lambda$ parameter assuming the 
Gaussian model, a linear structure for the regression and independent zero-mean errors. 
The horizontal line represents the value of $\lambda_c$ corresponding to white noise (uncorrelated stochastic dynamics).
The list of names for each map is the same given in Sec. \ref{sec:Chaos}.
Full circles (blue) are in agreement with Lacassa and Toral \cite{Lacasa2010} proposal rule.
Empty circles (red) not.
}
\end{figure}

\begin{figure} [ht!]
\noindent
\centering
\includegraphics[width=6.5in]{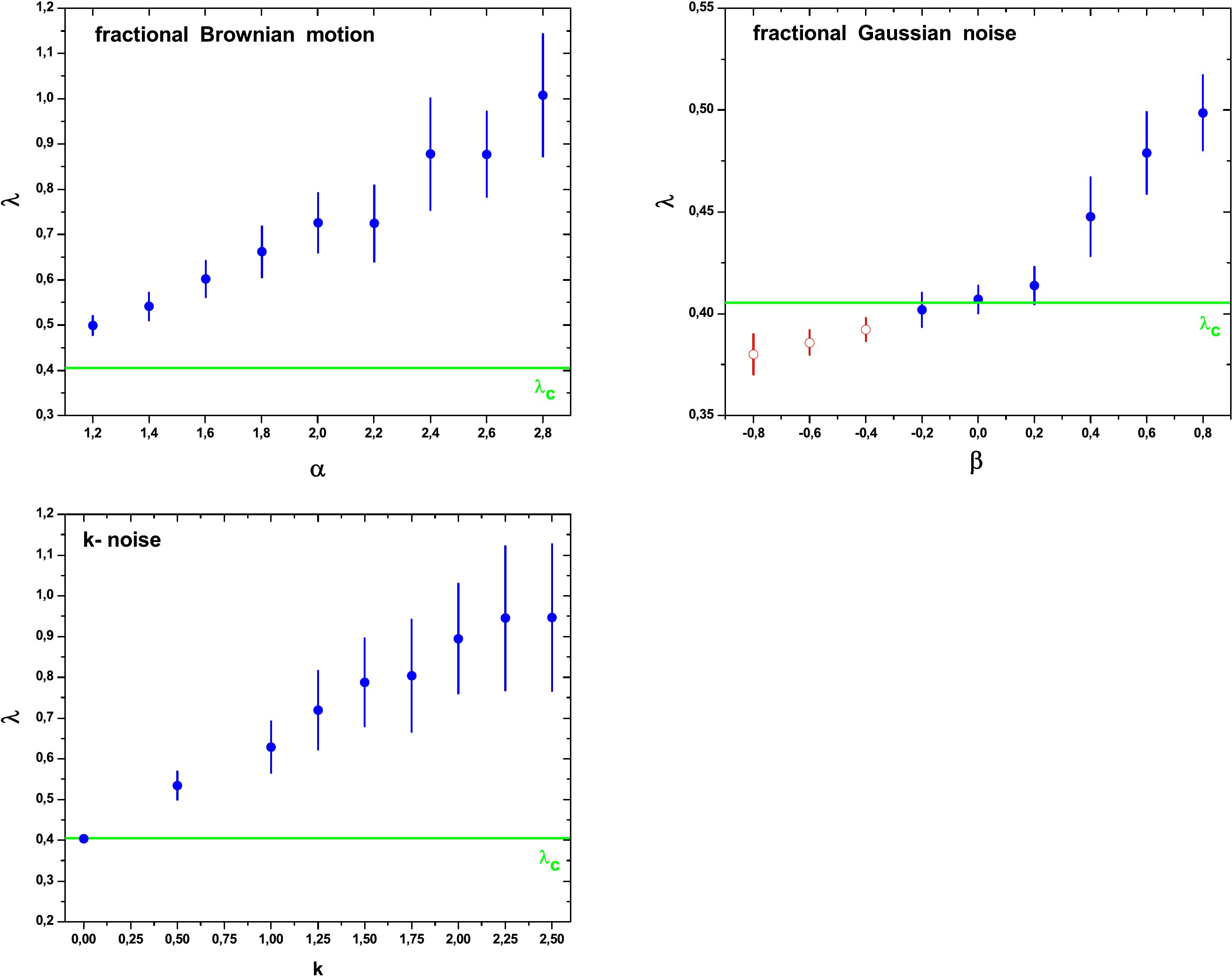}
\caption{ 
\label{fig:HVG-lambda-int-2} 
Parameter $\lambda$  values of  HVG-PDF $P(\kappa)$ for fBm, fGn and noise with $f^{-k}$ power spectrum time series
with total length of $N=10^5$ data. 
The $\lambda$ values were obtained following the methodology proposed by Lacasa {\it et al.}:  
from the graph $\ln[P(\kappa)]$ versus $\kappa$, the $\lambda$ parameter was computed by adjusting using the least
square method, a straight line being $\lambda$ its slope. 
The linear scaling region considered in all cases is $3 \leq \kappa \leq 20$, or $3 \leq \kappa \leq \kappa_{\max}$ 
(if $\kappa_{\max} \leq 20$).
Symmetric confidence intervals at the $95 \%$ confidence level were obtained for the $\lambda$ parameter assuming the 
Gaussian model, a linear structure for the regression and independent zero-mean errors. 
The horizontal line represents the value of $\lambda_c$ corresponding to white noise (uncorrelated stochastic dynamics)
Full circles (blue) are in agreement with Lacassa and Toral \cite{Lacasa2010} proposal rule.
Empty circles (red) not.}
\end{figure}

In the same fashion, we have computed the node degree distribution $P(\kappa)$ of the HVG for all the systems described in 
Section~\ref{sec:Chaos-Ruidos} for series of $10^5$ data length. 
Symmetric confidence intervals at the $95\%$ confidence level were obtained assuming the Gaussian model, a linear structure 
for the regression and independent zero-mean errors. The results are shown in Figures \ref{fig:HVG-lambda-int-1} and \ref{fig:HVG-lambda-int-2} 
for all studied chaotic and stochastic dynamics, respectively. It is possible to see from these figures that several chaotic and stochastic systems 
follow the above mentioned rule; however, we have found others that do not. Examples of chaotic maps for which $\lambda$ is larger than $\lambda_c$ 
(see Fig. \ref{fig:HVG-lambda-int-1} open circles) correspond to:
(5) cubic map,
(9) Pinchers map,
(10) Spence map,
(11) sine-circle map,
(14) delay logistic map,
(15) Tinkerbell map (X),
(16) Burger's map (Y),
(17) Holmes cubic map,
(19) Ikeda map (Y)
(21) discrete predator-prey map (Y),
(23) H\'enon area-preserving quadratic map,
(26) chaotic web map,
(27) Lorenz three-dimensional chaotic map,
(see also Table S1 in supplementary material).
Stochastic processes for which  $\lambda$ is smaller than $\lambda_c$
(see Fig. \ref{fig:HVG-lambda-int-2} open circles) correspond to fGn with $-0.8 \leq \beta \leq -0.4$.

\begin{figure} [ht!]
\noindent
\includegraphics[width=6.5in]{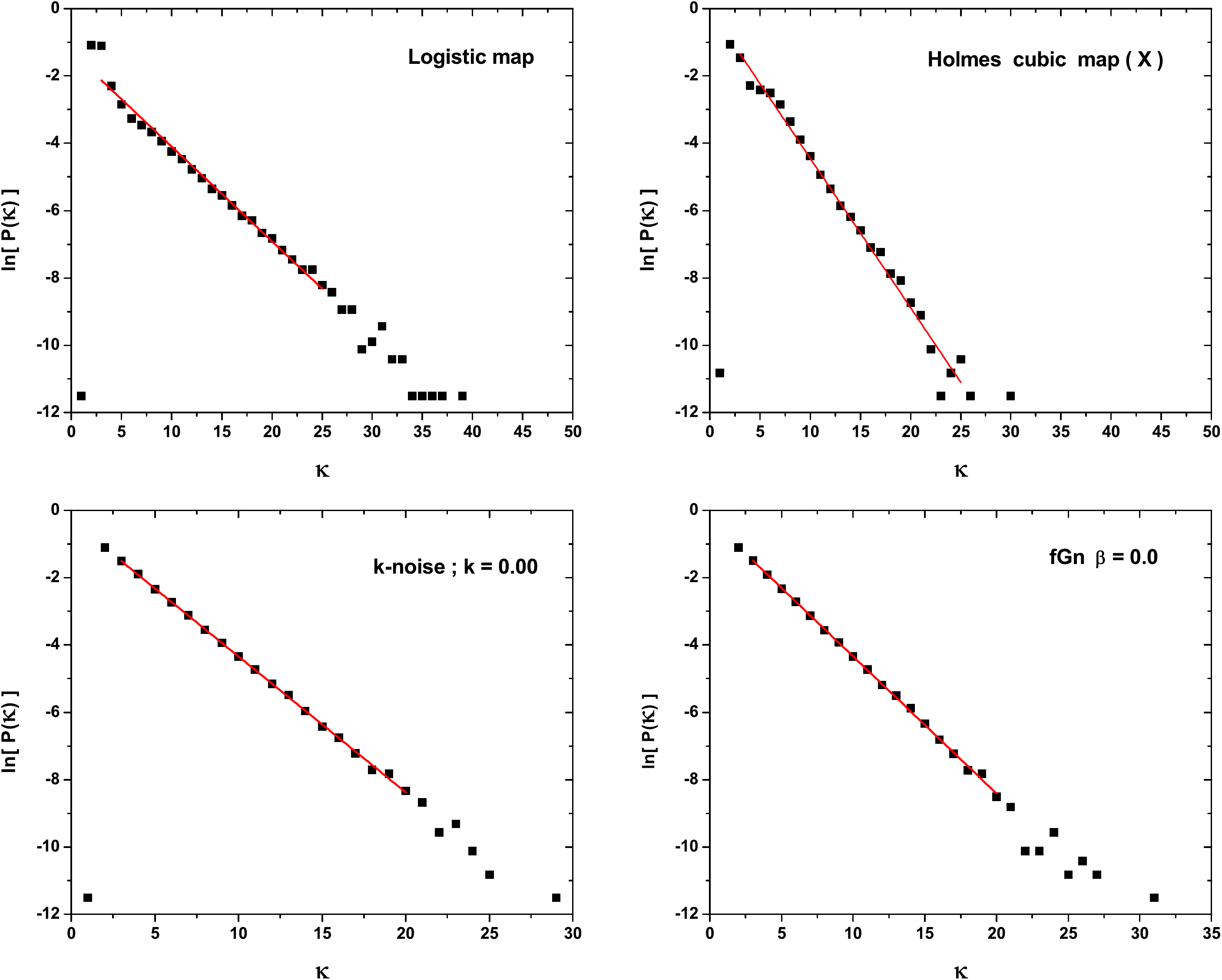}
\caption{
\label{fig:HVG-lambda-fiteo-bueno} 
$\lambda$-value determination: examples of analyzed dynamical systems where a good linear scaling region was found. 
For the Holmes cubic map ($X$), however, even having a good fitting, the  $\lambda$-value obtained is greater than 
$\lambda_c$ which  not satisfied the chaotic distinction suggested by Lacasa and Toral \cite{Lacasa2010}. 
In all cases, time series with $N=10^5$ are considered, and linear scaling regions are defined by $3 \leq \kappa \le 25$ 
for chaotic and $3 \leq \kappa \leq 20$ for stochastic time series.}
\end{figure}

\begin{figure} [ht!]
\centering
\includegraphics[width=3.5in]{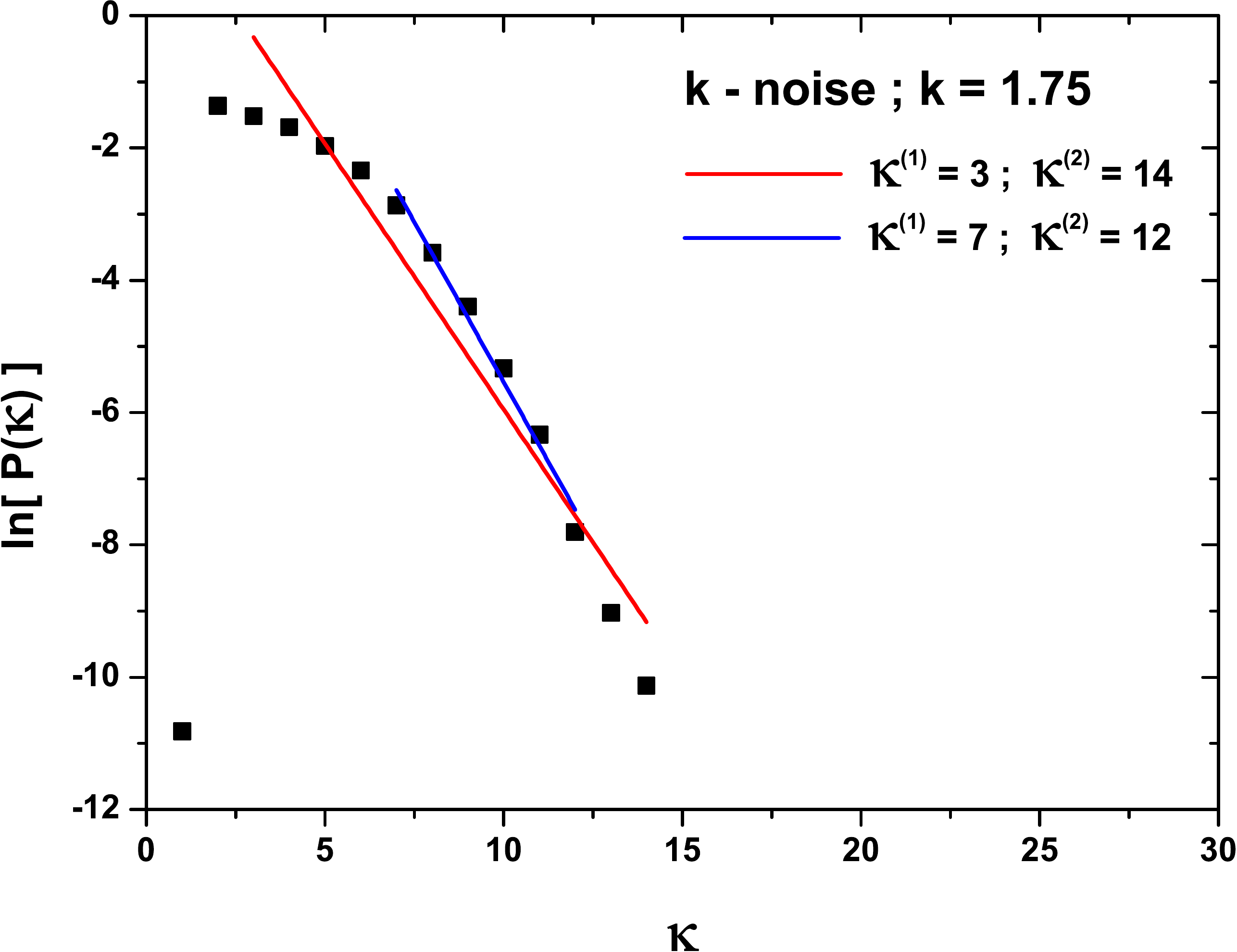}
\caption{ \label{fig:HVG-lambda-fiteo-malo-1}
$\lambda$-value determination in the case of time series generated by stochastic dynamics with
$f^{-k}$ power spectrum with $k= 1,75$.  Time series with $N=10^5$ data. 
Two different linear scaling zones: {\it a)\/}  $3 \leq \kappa \leq 14 $ given $ \lambda = 0.803~[0.666;0.941]$; and
{\it b)\/}  $7 \leq \kappa \leq 12$ given $ \lambda = 0.966~[0.797;1.136]$. Note that the slope of the straight line change significantly.}
\end{figure}

\begin{figure} [ht!]
\noindent
\centering
\includegraphics[width=6.5in]{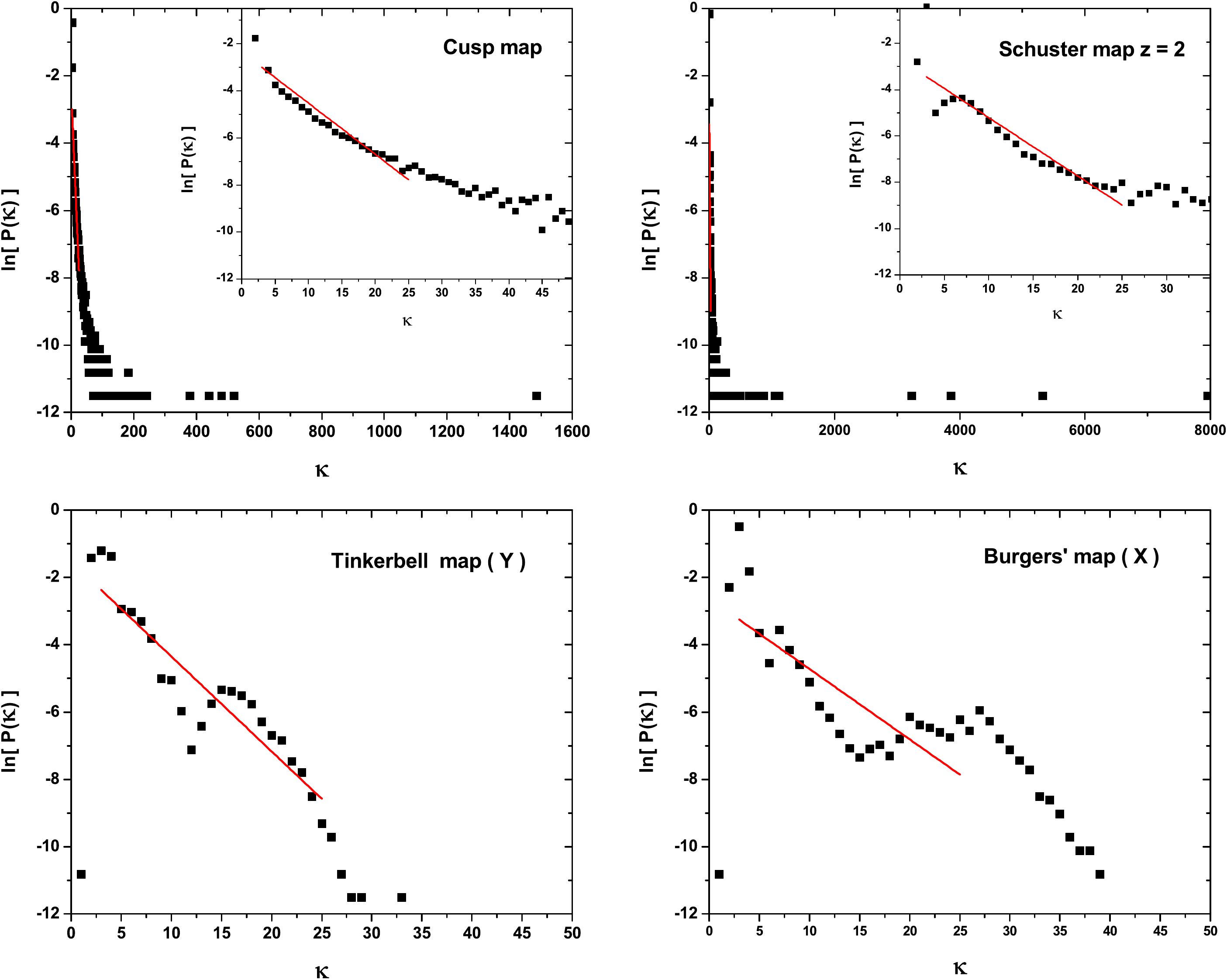}
\caption{
\label{fig:HVG-lambda-fiteo-malo-2}
Cases with bad $\lambda$-value determination: 
{\it a)\/} Cusp map and Schuster map with $z=2$, the associated HVG-PDF present heavy tail making difficult to define an 
unique linear scaling zone representative of all the data.
{\it b)\/} Tinkerbell map (Y) and the Burger's map (X) for which it is impossible  to define an unique linear scaling zone, 
and in consequence the hypothesis of an exponential behavior cannot be confirmed.
Time series with $N=10^5$ data are considered. 
}
\end{figure}

Some important issues to be discussed are:
\begin{description}
\item[Scaling zone:] Several systems present a well defined linear scaling region allowing a good linear fitting to obtain $\lambda$. 
Examples are the Logistic map, Holmes cubic map (X), a $k$-noise with $k=0$ and a fBm with $\beta=0$ presented in 
Figure~\ref{fig:HVG-lambda-fiteo-bueno}. However, we must point out that the fact of having a well scaling region does 
not guarantee the satisfaction of the $\lambda$ rule. 
See for instance the Holmes cubic map (X) that present a clear linear scaling region, however $\lambda = 0.443~[0.418;0.469] > \lambda_c$,
contradicting the hypothesis.
 
Another important point is the selection of the scaling zone, as the inclusion or exclusion of a few points in the extremes of the PDF may drastically change the $\lambda$ value. Figure~\ref{fig:HVG-lambda-fiteo-malo-1} shows the effect of selecting different scaling zones for a stochastic process with $f^{-1.75}$ PS . If the scaling zone is defined in the node degree interval $3 \leq \kappa \leq 14$, $\lambda = 0.803~[0.666;0.941]$, however, if the scaling zone is redefined for the interval $7 \leq \kappa \leq12$, $\lambda = 0.966~[0.797;1.136]$, which represent a variation of $16.3~\%$.

\item[Heavytailedness:] The definition of an unique linear scaling zone is a difficult task for systems with a heavy tailed PDF. 
Note that, when defining a scaling zone, important information contained in the tails may be lost. 
Examples of systems with heavy tailed PDFs are the Cusp and the Schuster maps (see Fig. \ref{fig:HVG-lambda-fiteo-malo-2}).

\item[Nonexponential behavior:] Some systems present PDFs with no linear scaling zone, in consequence, the hypothesis 
of an exponential behavior cannot be confirmed. See for instance, the Tinkerbell map (Y) and the Burger's map (X) in Fig. \ref{fig:HVG-lambda-fiteo-malo-2}.

\end{description}

In Table S1 readers find the $\lambda$ values with the corresponding confidence intervals, the coefficient of determination $R^2$,
as well all the corresponding plots for all the dynamical systems analyzed in this work (see Figures S4-S9).

\subsubsection{Skewness and kurtosis}
\label{sec:Skewness-kurtosis}

Given a one-dimensional probability distribution $f(x)$ with $x \in \Delta \subset {\mathbb R}$,
the usual spread measure is
the variance $V[f] =\int_{\Delta} ( x - \langle x \rangle)^2 dx$.
The variance measures the (quadratic) variability around the mean. This property makes the variance (or its square root, the standard deviation) 
particularly useful for smooth unimodal distributions.
Other interesting quantifiers based on higher moments order are the skewness (a third order moment measure) and the kurtosis (which depends on the fourth order moment).
The skewness measures the asymmetry, while the kurtosis describes the relative ``peakedness'' of the density with respect to the Gaussian law. Kurtosis is a sign of ``flattening'' or ``peakedness" of a distribution.

The usual skewness and kurtosis are of limited use and interpretability when dealing with asymmetric distributions, 
as is the case of the node degree distribution HVG-PDF $P(\kappa)$, which is always non-negative.
%It presents a node range variation $0 < \kappa \leq \kappa_{\max}$ where $\kappa_{\max}$ is a node cut-off from which $P(\kappa)=0,~\forall~ \kappa > \kappa_{\max}$ and could display other maximum values in between. 

%%%%%%%%%%%%%%%%%%%%%%%%%%%%%%%%%%%%%%%%%%%%%%%%%%%%%%%%%%%%%%%%%%%%%%%%%%%%%%%%%%%%%%%%%%%%%%%%%%%%%%%%%%%%%
%{\hfill\break
%\bf NOTA: QUIZAS HABRIA QUE MOSTRAR ALGUNA PDF PARA CAOTICOS + LOS SCHUSTERS (HAY QUE HACER LOS GRAFICOS !!!}
%{\hfill\break
%%%%%%%%%%%%%%%%%%%%%%%%%%%%%%%%%%%%%%%%%%%%%%%%%%%%%%%%%%%%%%%%%%%%%%%%%%%%%%%%%%%%%%%%%%%%%%%%%%%%%%%%%%%%%

Among the many alternatives available in the literature, for skewness and kurtosis  evaluation, Brys {\it et al.\/}~\cite{Brys2009} employ with success the information provided by the quantiles.
In particular, we will see that an alternative measure of kurtosis is  able to describe the different heavytailedness 
of the observed node degree distribution HVG-PDF.

Consider $\bm x = (x_1,\cdots,x_n)$ a sample of $n$ real values.
The sample quantile of order $0 < \eta < 1$ is  $q(\eta) = \min \{ x \in {\mathbb R} : \widehat{F} (x) \geq \eta \}$ 
and $\widehat{F}(x) = n^{-1} \# \{x_i : x_i \leq x\}$ is the sample cumulative distribution function also known as 
empirical function.
Quantile-based measures of skewness and kurtosis can be defined as

\begin{equation}
\label{eq:gamma1}
\widehat{\gamma_1}(\xi) ~=~ \frac{q(1-\xi)+q(\xi)-2q(1/2)}{q(1-\xi)-q(\xi)} \ , 
\end{equation}

and

\begin{equation}
\label{eq:gamma2}
\widehat{\gamma_2}({\xi,\varrho}) ~=~ \frac{q(1-\varrho)-q(\varrho)}{q(1-\xi)-q(\xi)}
\end{equation}
respectively, where $0<\xi<\varrho<1$ are arbitrary.
   
The values for two reference distributions were computed analytically with $\xi=1/10$ and $\varrho=1/100$.
For the standard exponential distribution with probability density function $P(\kappa) = e^{-\kappa}$, $\kappa>0$ 
they are:
\begin{equation}
\label{eq:gamma1B}
\gamma_1(1/10) ~=~ \frac{2\log(1/2)-\log(1/10)-\log(9/10)}{\log(9/10)-\log(1/10)} ~\approx~ 0.465 \ , 
\end{equation}
and
\begin{equation}
\label{eq:gamma2B}
\gamma_2(1/10,1/100) ~=~ \frac{\log(99/100)-\log(1/100)}{\log(9/10)-\log(1/10)} ~\approx~ 2.091 \ ,
\end{equation}
and for the node degree distribution under white noise, whose probability function is $P(k) = 3^{-1} (2/3)^{k-2}$, 
$k\geq2$ \cite{Lacasa2008,Lacasa2010}, they are $\gamma_1(1/10)=3/5=0.6$ and $\gamma_2(1/10,1/100) = 11/5=2.2$.

Table \ref{tab:SystemQuantifiers} shows the values of lambda ($\widehat\lambda$), skewness ($\widehat{\gamma_1}$) and kurtosis ($\widehat{\gamma_2}$) for several noises and chaotic maps.

\begin{figure} [ht!]
\noindent
\centering
\includegraphics[width=6.5in]{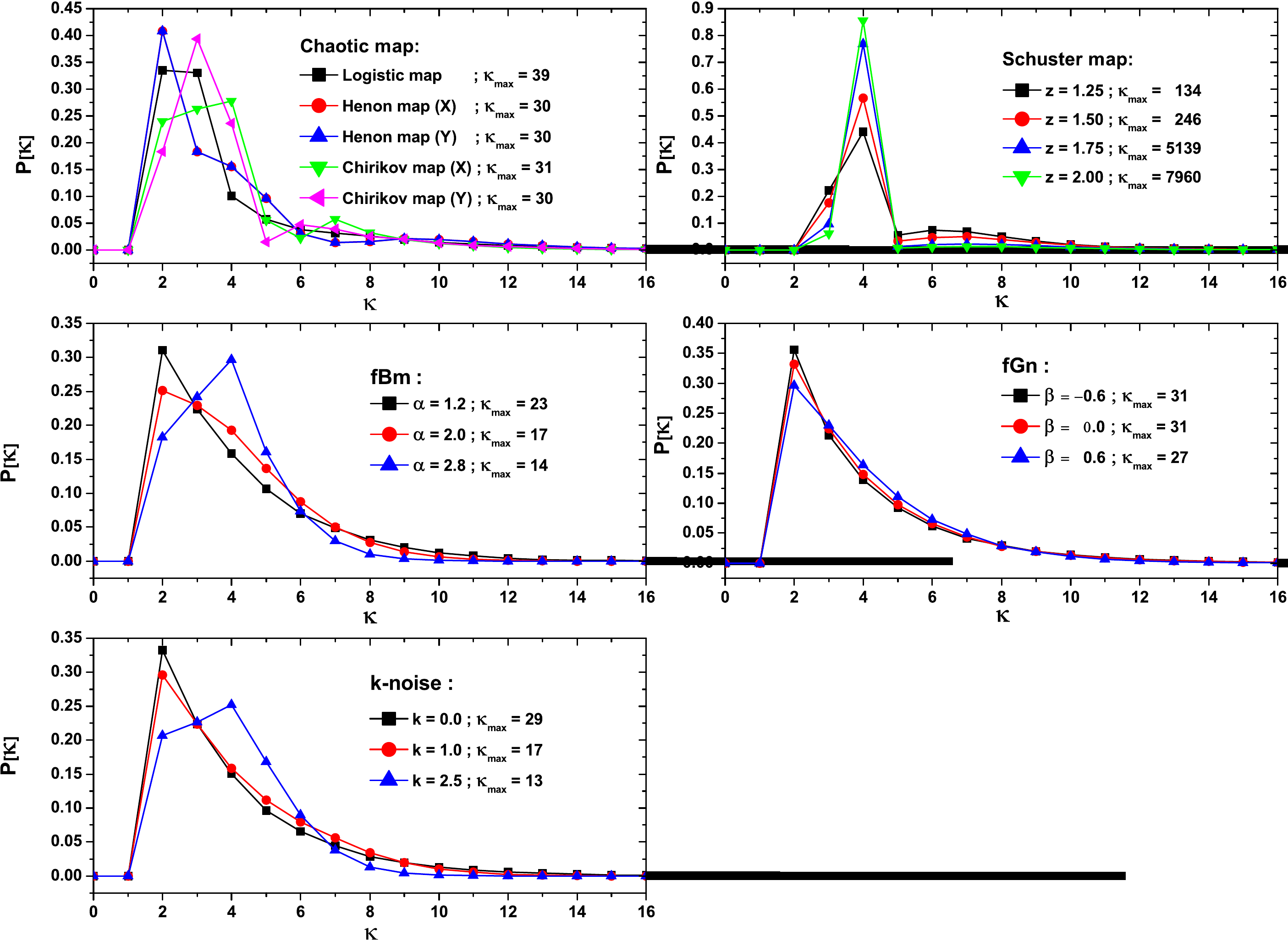}
\caption{ \label{fig:HVG-3PDF}  Examples of HVG-PDF for some chaotic and stochastic systems. Only $0 \leq \kappa \leq 16$  are displayed. Note that the corresponding cut-offs ($\kappa_{\max}$) are also shown. The length of the time series is $N=100,000$.}
\end{figure}

It is worth noticing that several chaotic maps present high kurtosis values indicating a heavy tailed PDF, showing the importance of using a quantifier that considers the entire available data. Fig. \ref{fig:HVG-3PDF} displays examples of HVG-PDF of several chaotic and stochastic systems. Note that, for some
systems, the HVG-PDFs do not present an exponential behavior.
Readers can find the results for all the systems considered in Table S1 of the supplementary material.

%Interesting, for some of the chaotic analyzed systems, the corresponding HVG-PDF present long tails, like the case  
%of Schuster's maps. 
\subsection{The Shannon-Fisher information plane}
\label{sec:Shannon-Fisher}

To avoid the subjectivity of choosing the scaling zone in which the parameter $\lambda$ is computed and, consequently, the 
sensitivity of this methodology, we propose a tool in which no information is lost, as the entire PDF is used and the relation 
between global and local features of the systems is captured. The Shannon-Fisher information plane 
(${\mathcal S} \times {\mathcal F}$) firstly introduced by Vignat and  Bercher \cite{Vignat2003} is a planar 
representation in which the horizontal and vertical axes are functionals of the pertinent probability distribution, 
namely, the  Shannon Entropy  ${\mathcal S}$ and the Fisher Information measure ${\mathcal F}$, respectively. 
This tool is a convenient way to represent in the same information plane global and local aspects of the PDFs associated to 
the studied system.  
In this work the PDFs are obtained through the horizontal visibility graph methodology \cite{Lacasa2008}.
 
Given a continuous probability distribution function (PDF) $f(x)$ with $x \in \Delta \subset {\mathbb R}$ and 
$\int_{\Delta} f(x)~dx = 1$, its {\it Shannon Entropy\/}  \cite{Shannon1948} is
\begin{equation}
\label{shannon}
{\mathcal S}[f]~=~-\int_{\Delta}~f~\ln f~dx \ ,
\end{equation}
a measure of {\it ``global"\/} character that it is not too sensitive to strong changes in the distribution 
taking place on small regions of the support $\Delta$. 

Such is not the case with {\it Fisher's Information Measure\/} (FIM) $\mathcal F$ \cite{Fisher1922,Frieden1998,Frieden2004}, 
which constitutes a measure of the gradient content of the distribution $f(x)$, thus being quite sensitive even 
to tiny localized perturbations. 
It reads
\begin{equation}
\label{fisher}
{\mathcal F}[f]~=~\int_{\Delta}~ { {1} \over {f(x)} } \left[ { {df(x)} \over {dx} }\right]^2 ~dx
    ~=~4 \int_{\Delta}~\left[ { {d \psi(x)} \over {dx} }\right]^2 \ .
\end{equation}
FIM can be variously interpreted as a measure of the ability to estimate a parameter, as the amount of information that 
can be extracted from a set of measurements, and also as a measure of the state of disorder of a system  or phenomenon 
\cite{Frieden2004}.
In the previous definition of FIM (Eq. (\ref{fisher})) the division by $f(x)$ is not convenient if $f(x)$ becomes too small to be adequately computed.
Such issue is avoided using probability amplitudes $\sqrt{f}$  \cite{Frieden1998,Frieden2004}.
The gradient operator significantly influences the contribution of minute local $f-$variations to FIM's value.
Accordingly, this quantifier is called a {\it ``local"\/} one \cite{Frieden2004}.

% We avoid this if we work with a real probability amplitudes $f(x)= \psi^{2}(x)$ \cite{Frieden1998,Frieden2004}, which is a  simpler form (no divisors) and shows that $\mathcal F$ simply measures the gradient content in $\psi(x)$.

Let now $P=\{p_i;~i=1,\dots, N\}$  be a  discrete probability distribution, with $N$ the number of possible states of 
the system under study. 
The concomitant  problem of information-loss due to discretization has been thoroughly studied (see, for instance, 
\cite{Zografos1986,Pardo1994,Madiman2007}, and references therein) and, in particular, it entails the loss of FIM's 
shift-invariance, which is of no importance for our present purposes \cite{Olivares2012A,Olivares2012B}.
In the discrete case,  we define a ``normalized" Shannon entropy as
\begin{equation}
\label{shannon-disc}
{\mathit H}[P]~=~ {\mathcal S}[P]  / S_{\max} ~=~\left\{-\sum_{i=1}^{N}~p_i~\ln p_i \right\} /  S_{\max} \ ,
\end{equation}
where the denominator  $S_{\max} = S[P_e] = \ln N$ is the Shannon entropy attained by a uniform probability distribution 
$P_e = \{p_i =1/N,~ \forall i = 1, \dots, N\}$.
For the FIM we take the expression in terms of real probability amplitudes as starting point, then a discrete normalized 
FIM convenient for our present purposes, is given by
\begin{equation}
\label{Fisher-disc}
{\mathcal F}[P]~=~F_0~\sum_{i=1}^{N-1}~[(p_{i+1})^{1/2} - (p_{i})^{1/2}]^2 \ .
\end{equation}
It has been extensively discussed that this discretization is the best behaved in a 
discrete environment \cite{Jesus2009}. Here the normalization constant $F_0$ reads
\begin{equation}
\label{F0}
F_0~=~\left\{
       \begin{array}{cl}
                    1       &\qquad \mbox{if $p_{i^*} = 1$ for
                            $i^* = 1$ or $i^* = N$ and $p_{i}  = 0 ~\forall  i \neq i^*$} \\
                    1/2     &\qquad \mbox{otherwise} \ .
       \end{array}
\right.
%\ .
\end{equation}

\section{Results and discussion}
\label{sec:Results}
\begin{figure} [ht!]
\noindent
\centering
\includegraphics[width=6.5in]{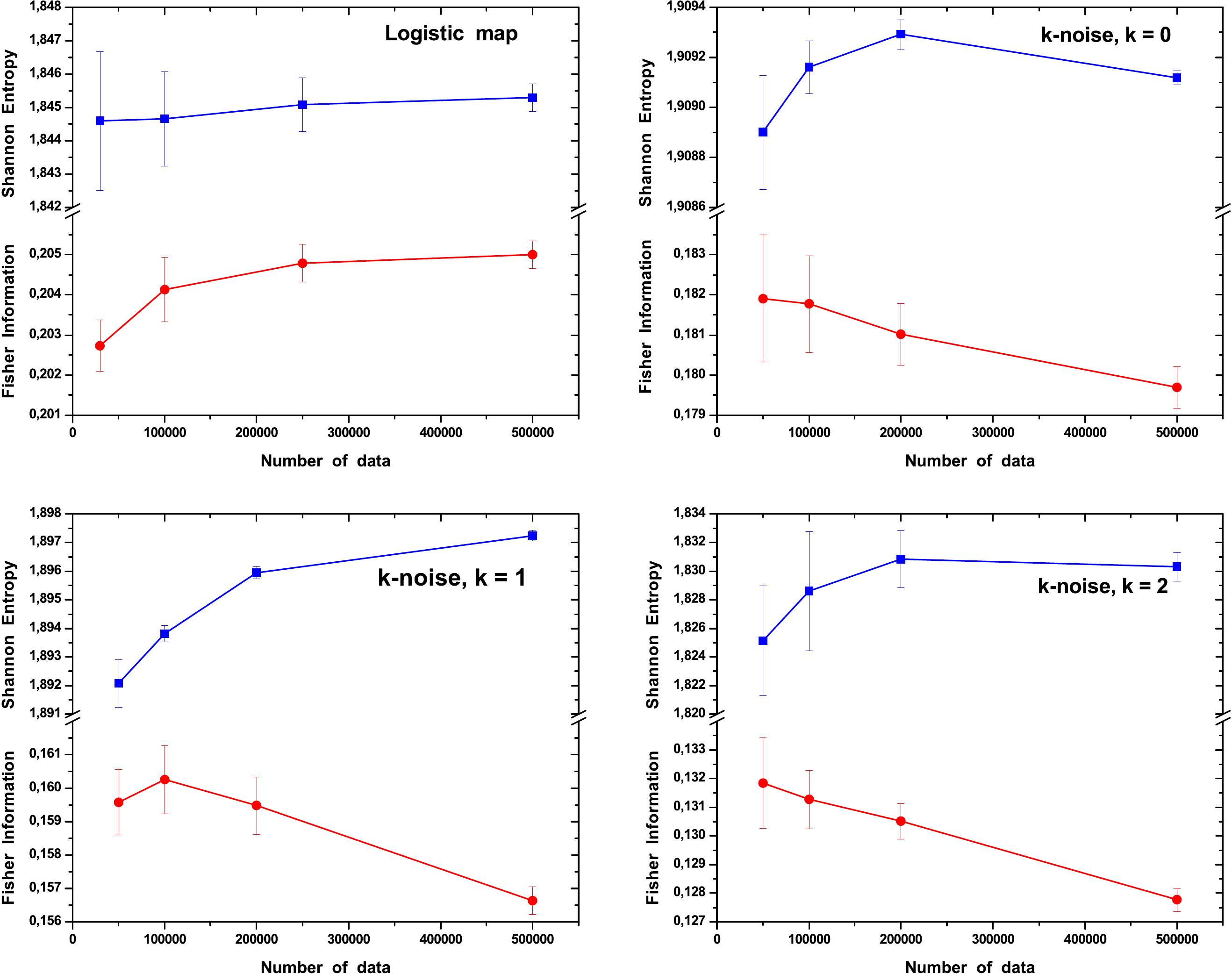}
\caption{
\label{fig:NVar} 
Study of the effect of the series length on the Information Theory quantifiers. 
The dynamical systems here considered are the Logistic map and noises with  $f^{-k}$ power spectrum, 
for $k=0,1$ and $2$.
}
\end{figure}

\begin{figure} [ht!]
\centering
\noindent
\includegraphics[width=6.5in]{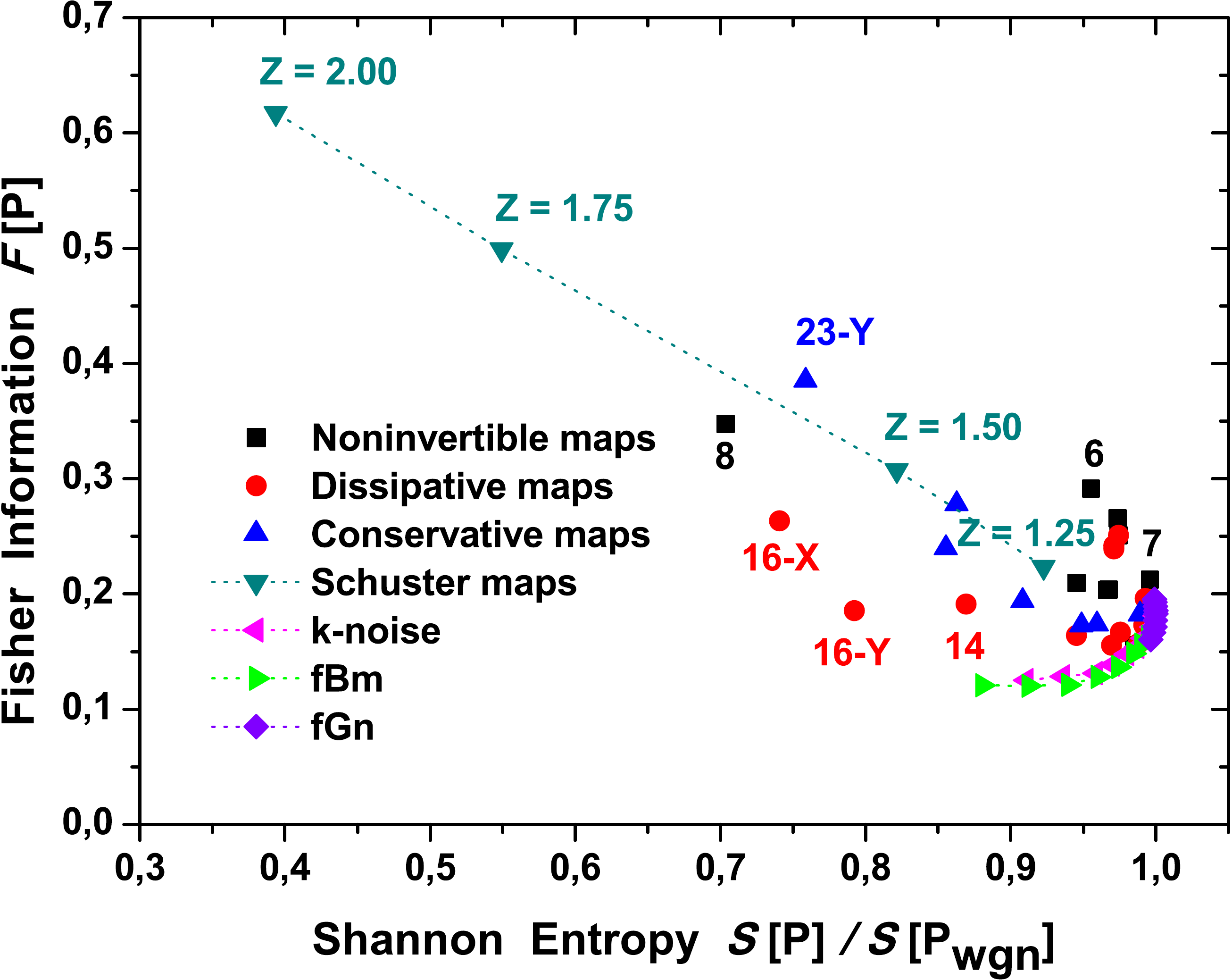}
\caption{
Representation on the Shannon-Fisher plane, ${\mathcal S} \times {\mathcal F}$, for all dynamical systems.
The quantifiers were evaluated with the HVG-PDF from time series length $10^5$.}
\label{fig:HVG-HxF-A}
\end{figure}

\begin{figure} [ht!]
\centering
\noindent
\includegraphics[width=6.5in]{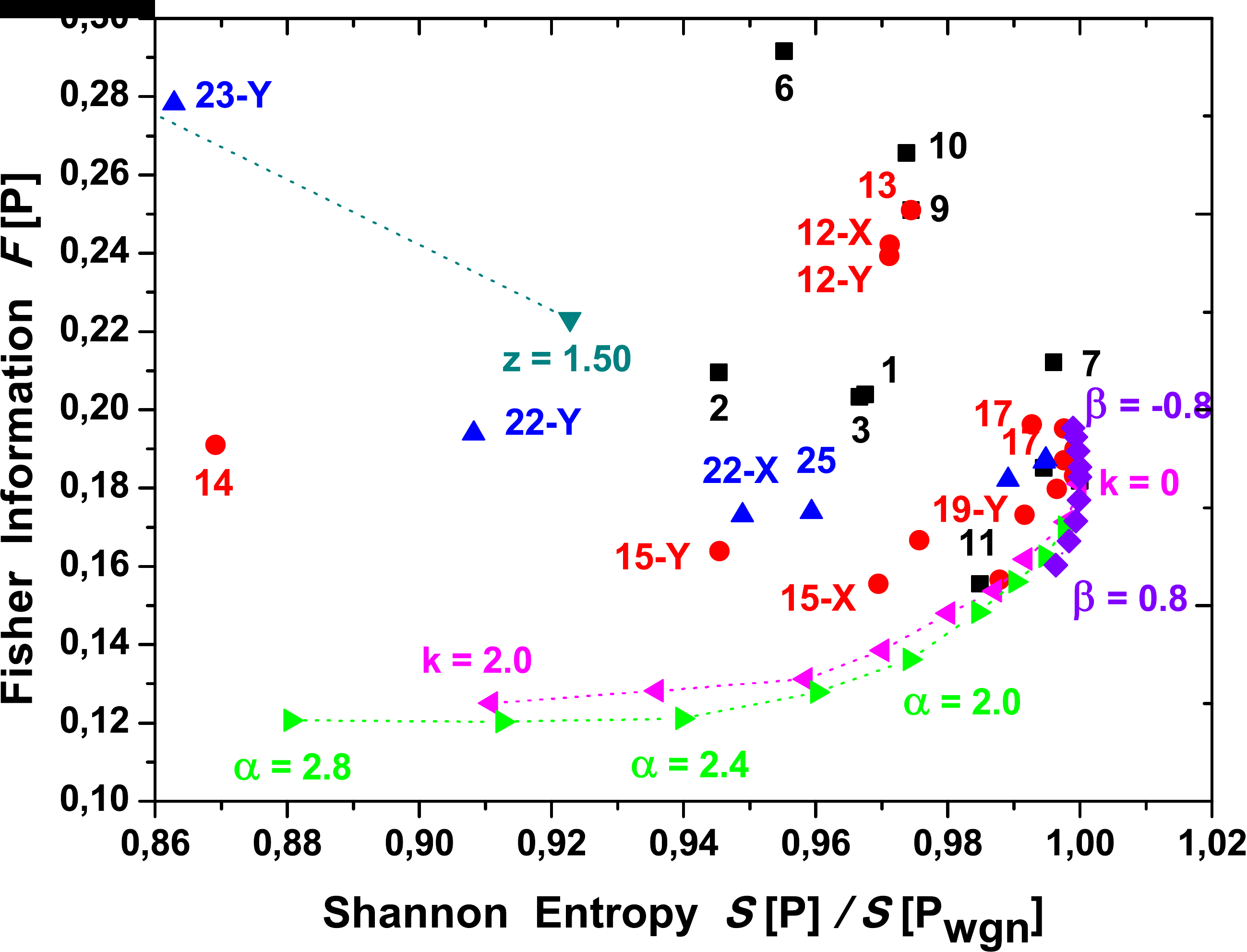}
\caption{\label{fig:HVG-HxF-zoom-1}
Shannon-Fisher plane, ${\mathcal S} \times {\mathcal F}$ zoom, see Fig. \ref{fig:HVG-HxF-A} 
}
\end{figure}

\begin{figure} [ht!]
\centering
\noindent
\includegraphics[width=6.5in]{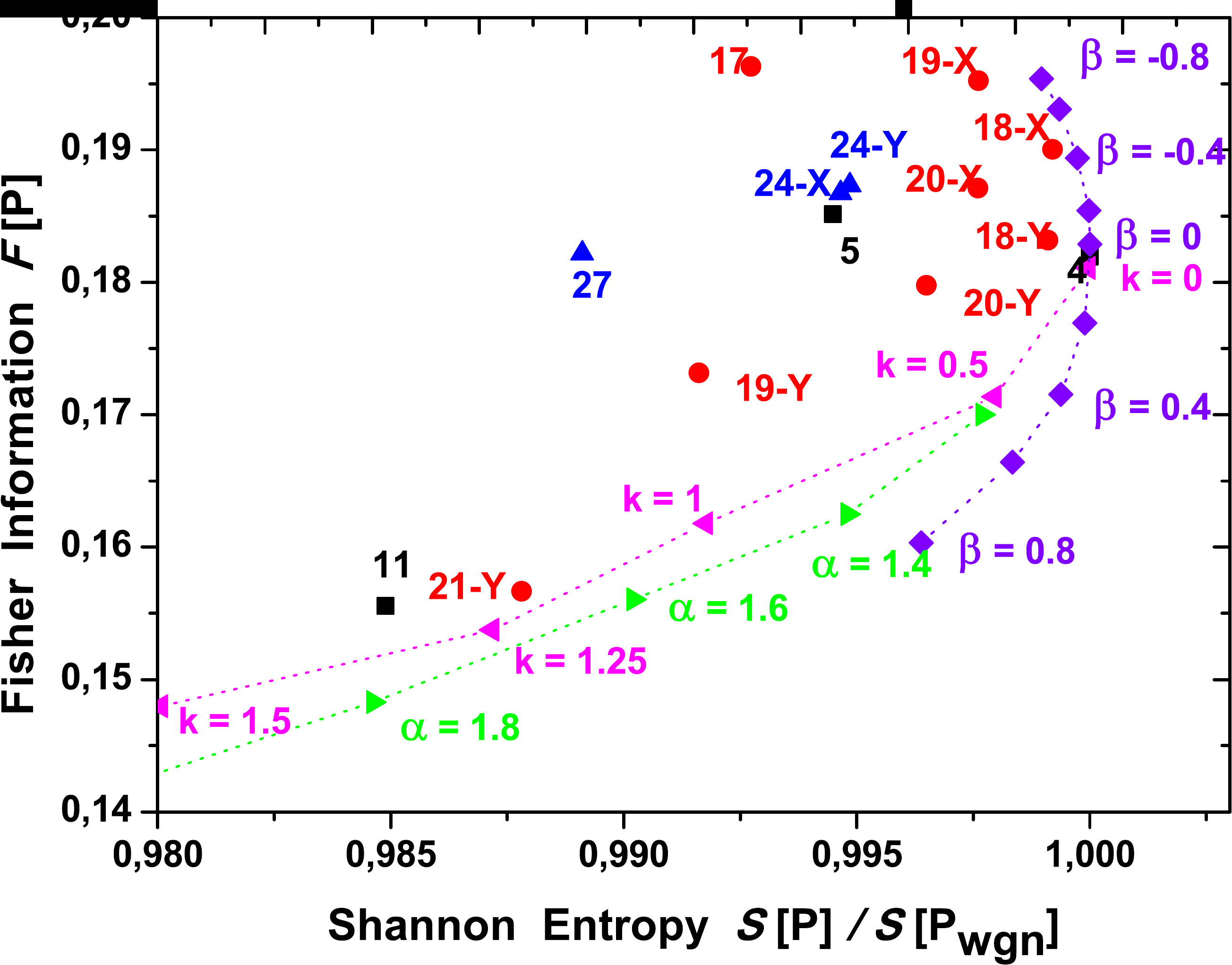}
\caption{
\label{fig:HVG-HxF-zoom-2}
Shannon-Fisher plane, ${\mathcal S} \times {\mathcal F}$ zoom, see Fig. \ref{fig:HVG-HxF-A} 
}
\end{figure}

In order to study the stability of the forthcoming results, we first analyze the dependency of the Information Theory quantifiers with the size of the time series. 
For this experiment we consider time series with different length sizes, varying from $30,000$ to $500,000$ values. 
As it can be seen in Figure~\ref{fig:NVar}, the Fisher Information and the Shannon entropy rapidly converge to 
stable values. For example, for the cases depicted in Figure~\ref{fig:NVar}, the order of magnitude of the percentage variations of the mean value for times series with $N=100,000$ and $N=500,000$ are between $10^{-4}$ and $10^{-5}$. For that reason all experiments consider time series with $100,000$ values. 
The Fisher Information and the Shannon Entropy are computed for all systems presented in Section \ref{sec:Chaos-Ruidos}. Results are depicted in Figures \ref{fig:HVG-HxF-A}, \ref{fig:HVG-HxF-zoom-1} and \ref{fig:HVG-HxF-zoom-2}. The Shannon entropy values are normalized with its maximum value for $N=100,000$, that corresponds to the entropy of the gaussian white noise (fBm for $\beta=0$, $\mathcal S_{[P_{wgn}]}$).
%Note that for bi-dimensional maps if one of the map coordinates has the same value than the other delay map coordinate (i.e.: delay logistic map), then no variation in the quantifiers evaluated with the HVG-PDF are expected.
One interesting observation resulting from this experiment is the fact that the Fisher Information 
(${\mathcal F}$) decreases with the strength of correlation in noises. 
The degree distribution corresponding to noises with lower correlation presents high peaks 
as well as long tails, almost flat for the white noise. 
As correlations get stronger, the peaks decrease and tails get shorter. 
For the uncorrelated situation (white noise), the strong contribution of the long and flat tail, 
even having the highest peak, makes the shape of the distribution more uniform. 
This effect can be seen in Figure \ref{fig:HVG-3PDF} 
as well as in Table S1 for noises with $f^{-k}$ power spectrum.

The Fisher Information is sensitive to small 
fluctuations. From Eq. (\ref{Fisher-disc}), it is possible to see that bigger differences in consecutive $p_i$ 
values of the distribution $P$, result in higher values of ${\mathcal F}$. 
In this case, the higher peaks in the degree distributions, that correspond to lower correlation 
values, represent the main contribution to ${\mathcal F}$.  
The extra terms present in the long tail, even contributing with small values, still increase 
the value of ${\mathcal F}$.
For that reason, the lowest value of ${\mathcal F}$ corresponds to the noise with strongest 
correlation structure (fBm with $\alpha= 2.8$), see Figure~\ref{fig:HVG-HxF-zoom-2}.

That is not the case of the Shannon entropy ${\mathcal S}$, which is
not sensitive to small fluctuations. The Shannon Entropy presents its highest value for noises with the smallest correlation
(white noises, gaussian and non-gaussian) and, as correlation structures get stronger, 
$\mathcal S$ decreases.

The planar localization in the Shannon-Fisher information plane ${\mathcal S} \times {\mathcal F}$, 
gives interesting information about the relation between the systems. 
Noises appear to be organized as a frontier, from which all chaotic maps concentrate. 
As it was previously shown, the frontier is stable regarding the size of the times series length.

Note that some chaotic maps are located nearby the noise ``frontier'' in the ${\mathcal S} \times {\mathcal F}$ plane (see Figure~\ref{fig:HVG-HxF-zoom-2}). These maps are: the linear congruential generator (4), the dissipative standard map (18), and the Sinai map (20).  
They present high ${\mathcal S}/{\mathcal S}_{wgn}$ values, $0.995 \leq {\mathcal S}/{\mathcal S}_{wgn} \leq 1$ and low $\mathcal F$ values, $0.17 \leq {\mathcal F} \leq 0.19$. This planar localization can be understood, as these maps present a stochastic like dynamical behavior when represented in a two dimensional plane. However, when represented in higher dimensional planes, planar structures appear denoting their chaotic behavior. 

The use of the ${\mathcal S} \times {\mathcal F}$ plane can shed light on the underlying system's structure.
For example, the Schuster maps display a linear behavior in the ${\mathcal S} \times {\mathcal F}$ 
plane when varying the $z$ parameter. Wider laminar regions ($z=2.00$) generate a greater number of nodes with lower 
degree values. At the same time nodes located in the extremes of a laminar region posses higher degree. 
As the parameter $z$ decreases, the laminar structures get thinner, reducing the number of nodes 
with higher degree value. This fact positioned the Schuster systems far from the frontier as $z$ increases as can be seen 
in Figure~\ref{fig:HVG-HxF-A}. 
In Table S1 in the supplementary material can be found a detailed description of the results. 

\section{Conclusions}
Time series (TS), temporal sequences of measurements or observations, are the basic elements  
for investigating natural phenomena. From TS one should judiciously extract information of dynamical systems. 
TS arising from chaotic systems share with those generated by stochastic processes several 
properties that make them very similar. Examples of these properties are: a wide-band power spectrum (PS), 
a delta-like autocorrelation function, and an irregular behavior of the measured signals.
As irregular and apparently unpredictable behavior is often observed in natural TS, the question that 
immediately emerges is whether the system is chaotic (low-dimensional deterministic) or stochastic.  
If one is able to show that the system is dominated by low-dimensional deterministic chaos, 
then only few (nonlinear and collective) modes are required to describe the pertinent dynamics 
\cite{Osborne1989}. If not, then, the complex behavior could be modeled by a system dominated 
by a very large number of excited modes which are in general better described by stochastic or 
statistical approaches. 

In the present contribution, we propose the use of the Horizontal Visibility Graph in combination 
with the Shannon entropy and the Fisher information measure as a methodology to study dynamical systems. 
Several chaotic and correlated noises were considered for an exhaustive analysis. 
The arrange of the results in the ${\mathcal S} \times {\mathcal F}$ plane, show that this novel tool is able to capture features
that reveal the nature governing the system.

Summarizing, we have presented extensive series of numerical evidence and have 
contrasted the characterizations of deterministic chaotic  and noisy-stochastic dynamics, 
as represented by time series of finite length. Surprisingly enough, one just has to look at the different planar locations of the two dynamical 
regimes. The planar location is able to tell us whether we deal with chaotic or stochastic time series.

% You may title this section "Methods" or "Models". 
% "Models" is not a valid title for PLoS ONE authors. However, PLoS ONE
% authors may use "Analysis" 
%\section*{Analysis}

% Do NOT remove this, even if you are not including acknowledgments
\section*{Acknowledgments}
%This research has been partially supported by 
M.G.R. acknowledges support from CNPq and FAPEMIG, Brazil.
L.C.C. acknowledges support from CNPq, Brazil.  
O.A.R. acknowledges support from Consejo Nacional de Investigaciones Cient\'{\i}ficas y T\'ecnicas (CONICET),
Argentina, and  FAPEAL, Brazil.
B. Amin Gon\c{c}alves acknowledges support from CAPES, Brazil.
A.C.F. acknowledges support from CNPq and FAPEAL, Brazil.

%\section*{References}
% The bibtex filename
%\bibliography{template}

%\section*{Figure Legends}
%\begin{figure}[!ht]
%\begin{center}
%%\includegraphics[width=4in]{figure_name.2}
%\end{center}
%\caption{
%{\bf Bold the first sentence.}  Rest of figure 2  caption.  Caption 
%should be left justified, as specified by the options to the caption 
%package.
%}
%\label{Figure_label}
%\end{figure}

%\section*{Tables}
%\begin{table}[!ht]
%\caption{
%\bf{Table title}}
%\begin{tabular}{|c|c|c|}
%table information
%\end{tabular}
%\begin{flushleft}Table caption
%\end{flushleft}
%\label{tab:label}
% \end{table}

\vfill\eject
\section*{Tables}
 
%%%%%%%%%%%%%%%%%%%%%%%%%%%%%%%%%%%%%%%%%%%%%%%%%%%%%%%%%%
%%%%%%%%%%%  TALA 1 : HVG-PDF Skewnes & Kurtosis
%%%%%%%%%%%%%%%%%%%%%%%%%%%%%%%%%%%%%%%%%%%%%%%%%%%%%%%%%%
%\newpage
%\setcounter{figure}{1}
\begin{table}[hbt]
\caption{
Dynamical systems and their statistical quantifiers skewness ($\widehat{\gamma_1}$), kurtosis ($\widehat{\gamma_2})$
evaluated for $\xi=1/10$ and $\varrho=1/100$.
$\widehat\lambda$ is the obtained parameter value (with confidence value interval) of exponential functional form 
proposed by Lacasa and Toral for the HVG-PDF \cite{Lacasa2010}).
Time series with $N=10^5$ data are considered.
}
\label{tab:SystemQuantifiers}
\centering
\begin{tabular}{lccc}\toprule
System 			& $\widehat\lambda$  	        & $\widehat{\gamma_1}(\xi)$ 	        & $\widehat{\gamma_2}({\xi,\varrho})$ \\ \midrule
Exponential 		& $1$ 				& $0.465$ 				& $2.091$\\
White Noise 		& $\log 3/2\approx 0.405$       & $0.600$ 				& $2.200$ \\ \midrule
$k=0$ Noise 		& $0.404~[0.398;0.409]$ 	& $0.600$ 				& $2.200$ \\
$\alpha=2$ fBm  	& $0.726~[0.660;0.791]$ 	& $0.200$ 				& $1.600$ \\
$\beta=0$ fGn 		& $0.407~[0.400;0.414]$ 	& $0.600 $ 				& $2.200$  \\
Logistic map 		& $0.281~[0.264;0.298]$ 	& $0.666$ 			        & $2.333$ \\
Cusp map  		& $0.217~[0.175;0.259]$ 	& $0.500$ 				& $5.250$ \\
$z=1.25$ Schuster  	& $0.387~[0.362;0.412]$ 	& $0.600 $ 				& $2$ \\
$z=2$ Schuster  	& $0.252~[0.196;0.308]$  & $-1$ 					& $13$ \\
\bottomrule
\end{tabular}
\end{table}

\vfill\eject
\section*{Figure Legends}

%%%%%%%%%%%%%%%%%%%%%%%%%%%%%%%%%%%%%%%%%%%%%%%%%%%%%%%%%%
%%%%%%%%%%%  FIGURA 1 : HVG esquema + P(k) tabla
%%%%%%%%%%%%%%%%%%%%%%%%%%%%%%%%%%%%%%%%%%%%%%%%%%%%%%%%%%
%\newpage
%\setcounter{figure}{1}

%%%%%%%%%%%%%%%%%%%%%%%%%%%%%%%%%%%%%%%%%%%%%%%%%%%%%%%%%%
%%%%%%%%%%%  FIGURA 2 : Lambda - chaotic dynamics
%%%%%%%%%%%%%%%%%%%%%%%%%%%%%%%%%%%%%%%%%%%%%%%%%%%%%%%%%%
%\newpage
%\setcounter{figure}{1}

%%%%%%%%%%%%%%%%%%%%%%%%%%%%%%%%%%%%%%%%%%%%%%%%%%%%%%%%%%
%%%%%%%%%%%  FIGURA 3 : Lambda - stochastics dynamics
%%%%%%%%%%%%%%%%%%%%%%%%%%%%%%%%%%%%%%%%%%%%%%%%%%%%%%%%%%
%\newpage
%\setcounter{figure}{1}

%%%%%%%%%%%%%%%%%%%%%%%%%%%%%%%%%%%%%%%%%%%%%%%%%%%%%%%%%%
%%%%%%%%%%%  FIGURA 4 : Lambda - fiteo-bueno
%%%%%%%%%%%%%%%%%%%%%%%%%%%%%%%%%%%%%%%%%%%%%%%%%%%%%%%%%%
%\newpage
%\setcounter{figure}{1}

%%%%%%%%%%%%%%%%%%%%%%%%%%%%%%%%%%%%%%%%%%%%%%%%%%%%%%%%%%
%%%%%%%%%%%  FIGURA 5 : Lambda - fiteo-malo-1
%%%%%%%%%%%%%%%%%%%%%%%%%%%%%%%%%%%%%%%%%%%%%%%%%%%%%%%%%%
%\newpage
%\setcounter{figure}{1}

%%%%%%%%%%%%%%%%%%%%%%%%%%%%%%%%%%%%%%%%%%%%%%%%%%%%%%%%%%
%%%%%%%%%%%  FIGURA 6 : Lambda - fiteo-malo-2
%%%%%%%%%%%%%%%%%%%%%%%%%%%%%%%%%%%%%%%%%%%%%%%%%%%%%%%%%%
%\newpage
%\setcounter{figure}{1}

%%%%%%%%%%%%%%%%%%%%%%%%%%%%%%%%%%%%%%%%%%%%%%%%%%%%%%%%%%
%%%%%%%%%%%  FIGURA 7 : HVG-PDF ejemplos
%%%%%%%%%%%%%%%%%%%%%%%%%%%%%%%%%%%%%%%%%%%%%%%%%%%%%%%%%%
%\newpage
%\setcounter{figure}{1}

%%%%%%%%%%%%%%%%%%%%%%%%%%%%%%%%%%%%%%%%%%%%%%%%%%%%%%%%%%
%%%%%%%%%%%  FIGURA 8 : HVG-N-variable
%%%%%%%%%%%%%%%%%%%%%%%%%%%%%%%%%%%%%%%%%%%%%%%%%%%%%%%%%%
%\newpage
%\setcounter{figure}{1}

%%%%%%%%%%%%%%%%%%%%%%%%%%%%%%%%%%%%%%%%%%%%%%%%%%%%%%%%%%
%%%%%%%%%%%  FIGURA 9 : HVG plano HxF
%%%%%%%%%%%%%%%%%%%%%%%%%%%%%%%%%%%%%%%%%%%%%%%%%%%%%%%%%%
%\newpage
%\setcounter{figure}{7}

%%%%%%%%%%%%%%%%%%%%%%%%%%%%%%%%%%%%%%%%%%%%%%%%%%%%%%%%%%
%%%%%%%%%%%  FIGURA 10 : HVG plano HxF - zoom 1
%%%%%%%%%%%%%%%%%%%%%%%%%%%%%%%%%%%%%%%%%%%%%%%%%%%%%%%%%%
%\newpage
%\setcounter{figure}{7}

%%%%%%%%%%%%%%%%%%%%%%%%%%%%%%%%%%%%%%%%%%%%%%%%%%%%%%%%%%
%%%%%%%%%%%  FIGURA 11 : HVG plano HxF - zoom 2
%%%%%%%%%%%%%%%%%%%%%%%%%%%%%%%%%%%%%%%%%%%%%%%%%%%%%%%%%%
%\newpage
%\setcounter{figure}{7}

\end{document}